\documentclass[10pt,twocolumn,letterpaper]{article}

\usepackage[pagenumbers]{cvpr} 

\usepackage[accsupp]{axessibility}  

\usepackage{soul}
\usepackage{color, xcolor}

\soulregister\cite7 
\soulregister\citep7 
\soulregister\citet7 
\soulregister\ref7 
\soulregister\pageref7 

\usepackage{graphicx}
\usepackage{amsmath}
\usepackage{amssymb}
\usepackage{booktabs}
\usepackage{multirow}
\usepackage{bm}
\usepackage[detect-all]{siunitx}
\usepackage[pagebackref,breaklinks,colorlinks]{hyperref}

\usepackage[capitalize]{cleveref}
\crefname{section}{Sec.}{Secs.}
\Crefname{section}{Section}{Sections}
\Crefname{table}{Table}{Tables}
\crefname{table}{Tab.}{Tabs.}

\def\cvprPaperID{1560} 
\def\confName{CVPR}
\def\confYear{2022}

\begin{document}

\newtheorem{theorem}{\textbf{Theorem}}
\newtheorem{definition}{\textbf{Definition}}
\newtheorem{proof}{\textbf{Proof}}

\title{Node Representation Learning in Graph via Node-to-Neighbourhood Mutual Information Maximization}

\author{Wei Dong$^{1}$,~ Junsheng Wu$^{2}$\footnotemark[1],~ Yi Luo$^{2}$,~ Zongyuan Ge$^{3}$,~ Peng Wang$^{4}$\thanks{Wei Dong's email is dw156@mail.nwpu.edu.cn. Corresponding authors: Peng Wang (pengw@uow.edu.au) and Junsheng Wu (wujunsheng@nwpu.edu.cn)} \\
$^{1}$School of Computer Science and Engineering, $^{2}$School of Software, Northwestern Polytechnical University \\
$^{3}$Monash University \\
$^{4}$University of Wollongong  \\
}
\maketitle
\begin{abstract}
The key towards learning informative node representations in graphs lies in how to gain contextual information from the neighbourhood. In this work, we present a simple-yet-effective self-supervised node representation learning strategy via directly maximizing the mutual information between the hidden representations of nodes and their neighbourhood, which can be theoretically justified by its link to graph smoothing. Following InfoNCE, our framework is optimized via a surrogate contrastive loss, where the positive selection underpins the quality and efficiency of representation learning. To this end, we propose a topology-aware positive sampling strategy, which samples positives from the neighbourhood by considering the structural dependencies between nodes and thus enables positive selection upfront. In the extreme case when only one positive is sampled, we fully avoid expensive neighbourhood aggregation. Our methods achieve promising performance on various node classification datasets. It is also worth mentioning by applying our loss function to MLP based node encoders, our methods can be orders of faster than existing solutions. Our codes and supplementary materials are available at \url{https://github.com/dongwei156/n2n}.
\end{abstract}
\section{Introduction}\label{sec:intro}
Graph-structured data is ubiquitous because almost nothing in the world exists in isolation. As an effective graph modeling tool, Graph Neural Networks (GNNs) have gained increasing popularity in a wide range of domains such as computer vision~\cite{zhao2019multi,shi2019skeleton,qi20173d}, natural language processing~\cite{wu2021graph}, knowledge representation~\cite{cui2018survey}, social networks~\cite{hamilton2017inductive}, and molecular property prediction~\cite{wieder2020compact}, just name a few.

In this work, we focus on the node classification task in graphs where the key is to learn informative structure-aware node representations by gaining contextual information from the neighbourhood. This motivates a massive proliferation of message passing techniques in GNNs. Among these methods, a dominant idea follows an AGGREGATION-COMBINE-PREDICTION pipeline, where AGGREGATION step aggregates the neighbouring information into vectorized representations via various neighbourhood aggregators such as \emph{mean}~\cite{hamilton2017inductive}, \emph{max}~\cite{hamilton2017inductive}, \emph{attention}~\cite{velickovic2019graph}, and \emph{ensemble}~\cite{corso2020principal}, which are COMBINED with the node representations via sum or concatenation to realize neighbourhood information fusion. To model multi-hop message passing, the AGGREGATION and COMBINE operations tend to be repeated before the ultimate node representations are obtained for label prediction. In other words, the information exchange in this pipeline is driven by a node classification loss in the PREDICTION phase. Under the umbrella of supervised learning, this line of methods unifies node representation learning and classification. A potential problem is that they may suffer from the scalability issue due to the expensive labeling cost for large-scale graph data.

As a remedy to expensive human annotation, Self-Supervised Learning (SSL) has shown proven success in computer vision~\cite{chen2020simple,chen2021exploring} and natural language processing~\cite{devlin2019bert,brown2020language}, which, however, is less sufficiently explored in graph modeling. The key challenge thereof lies in how to design suitable \textit{pretext} task from non-Euclidean graph-structured data to learn informative node representations. Inheriting the idea from computer vision, many recent attempts approach graph-based SSL by designing topological graph augmentations to generate multi-view graphs for contrastive loss. Off-the-shelf GNNs~\cite{kipf2017semi,xu2019how} tend to be used as default options for node/graph encoding.

In this work, we propose a simple-yet-effective SSL alternative to learn node representations by adopting aggregation-free Multi-layer Perceptron (MLP) as node encoder and directly maximizing the mutual information between the hidden representations of nodes and their neighbourhood. Intuitively, by aligning the node representation to the neighbourhood representation, we can distill useful contextual information from the surrounding, analogous to knowledge distillation~\cite{park2019relational,cho2019efficacy}. Theoretically, the proposed Node-to-Neighbourhood (N2N) mutual information maximization essentially encourages graph smoothing based on a quantifiable graph smoothness metric. Following InfoNCE~\cite{oord2018representation}, the mutual information can be optimized by a surrogate contrastive loss, where the key boils down to positive sample definition and selection.

To further improve the efficiency and scalability of our N2N network as well as ensuring the quality of selected positives, we propose a Topology-Aware Positive Sampling (TAPS) strategy, which samples positives for a node from the neighbourhood by considering the structural dependencies between nodes. This enables us to select the positives upfront. In the extreme case when only one positive is used for contrastive learning, we can avoid the time-consuming neighbourhood aggregation step, but still achieve promising node classification performance.

We conduct experiments on six graph-based node classification datasets and the results show the promising potential of the proposed node representation strategy. The contributions of this work can be summarized as follows:

\begin{itemize}
\item We propose a simple-yet-effective MLP-based self-supervised node representation learning strategy with the idea of maximizing the mutual information between nodes and surrounding neighbourhood. We reveal our N2N mutual information maximization strategy essentially encourages graph smoothing by resorting to a quantifiable smoothness metric.
\item The mutual information maximization problem is optimized by a surrogate contrastive loss. A scalable TAPS strategy is proposed which enables the positives to be selected upfront. In the case when only one positive is considered, we avoid expensive neighbourhood aggregation but still obtain satisfactory node classification performance.
\item Experiments on six graph-based node classification datasets show our methods not only achieve competitive performance but have other appealing properties as well. For example, our methods can be orders of faster than existing solutions.
\end{itemize}

\section{Related Work}\label{sec:rw}
In this section, we briefly review existing work on (1) Graph Neural Networks (GNNs) and (2) Graph Contrastive Learning (GCL).

\textbf{Graph Neural Networks.}~GNNs target to learn structure-aware node/graph representations based on the topology structure of the graph data~\cite{wu2019comprehensive}. A main focus thereof is to design effective message passing strategies to encourage information propagation in the graph. Early attempts~\cite{scarselli2008graph} learn node representations by following a recurrent fashion, where the node states are updated by applying a propagation function iteratively until equilibrium is reached. Inspired by convolutional neural networks that were proposed originally for grid-like topology, such as images, convolution-like propagation was introduced into graph data~\cite{defferrard2016convolutional}. As a prevailing variant of this line of work,  Graph Convolutional Network (GCN)~\cite{kipf2017semi} stacks a set of $1$-hop spectral filters~\cite{defferrard2016convolutional} and nonlinear activation function to learn node representations.
GCN ignites a wave of following-up work with the aim of improving the efficiency or effectiveness of information exchange in graphs. Simplifying Graph Convolutional Network (SGC)~\cite{wu2019simplifying} reduces the excessive complexity of GCN by removing the nonlinear activation function to obtain collapsed aggregation matrix. L2-GCN~\cite{you2020layer} simplifies the training complexity of GCN by using an efficient layer-wise training strategy. GraphSAGE-Mean~\cite{hamilton2017inductive} applies \emph{mean} aggregator to fixed number of randomly sampled neighbours to reduce the computational cost and adopts concatenation to merge the node and neighbourhood information. FastGCN~\cite{chen2018fastgcn} also adopts sampling strategy to reduce computational footprint of GCN by interpreting graph convolutions as integral transforms of embedding functions. Another line of work aims to design sophisticated neighbourhood aggregation strategies.
Graph Attention Network (GAT)~\cite{velickovic2019graph} stacks masked self-attention layers in order that the nodes can adaptively attend their neighbours. The use of multiple aggregators (\emph{mean, sum, max, normalized mean, scaling,} etc) is also proven to be able to benefit the expressive power of GNNs either empirically~\cite{dehmamy2019understanding} or theoretically~\cite{corso2020principal}. Parallelly, efforts are also dedicated towards theoretical understanding about the properties of various GNN models. For example, in~\cite{xu2019how}, the authors present a theoretical framework for analyzing the representational power of GNNs based on Weisfeiler-Lehman (WL) graph isomorphism test. In~\cite{hou2019measuring}, CS-GNN is designed to understand and improve the use of graph information in GNNs by leveraging feature and label based smoothness metrics.

The aforementioned GNNs are under the umbrella of supervised learning, where a predominant training pipeline is AGGREGATION-COMBINE-PREDICTION, which unifies node representation learning and node classification in an end-to-end fashion. The supervised GNNs show proven success in various applications by virtue of sufficient labeled data.

\textbf{Graph Contrastive Learning.}~With the aim of learning high-quality data representations without human annotation, SSL has shown very promising performance in computer vision and natural language processing. Motivated by such success, marrying SSL and GNNs is also gaining increasing attention, where the key boils down to designing effective \textit{pretext} tasks for graph-structured data. In this work, we focus mainly on Graph Contrastive Learning (GCL) as it is directly related to the proposed method.

One key in contrastive learning is how to define positive samples as it directly determines contrasting mode. The work in~\cite{you2020graph} directly performs graph-level contrastive learning, where two views of the same graph are obtained by performing augmentations such as dropping nodes, edge perturbation and attribute masking. The motivation behind is that the semantics in graphs are invariant to minor topological change. Another line of work is inspired by Deep InfoMax in computer vision~\cite{hjelm2018learning}, where the image representation is learned by maximizing the mutual information between the learned representation and the input image. By applying this mutual information maximization idea in graphs, several variants are derived. Deep Graph InfoMax (DGI)~\cite{velivckovic2018deep} aligns the global representation of a graph to its hidden node representation by contrasting against the node representations derived from a corrupted graph. To avoid the readout function and corrupted operation in~\cite{velivckovic2018deep}, Graphical Mutual Information (GMI)~\cite{peng2020graph} aligns the output node representation to the input sub-graph. The work in~\cite{hassani2020contrastive} learns node and graph representation by maximizing mutual information between node representations of one view and graph representations of another view obtained by graph diffusion. InfoGraph~\cite{sun2020infograph} works by taking graph representation and patch representation as pairs and determine whether they are from the same graph. Although our method also adopts mutual information maximization, our method differs from the aforementioned methods from several perspectives. Firstly, we avoid the operations such as graph augmentation, iterative aggregation and readout function by aligning nodes to their neighbourhood in the output layer of a MLP encoder. Secondly, instead of designing the \textit{pretext} task heuristically, our framework can be theoretically justified by its connection to graph smoothing. Finally, a topology-aware sampling strategy is proposed in our work that converts our framework into highly efficient node-to-node contrastive learning but still maintains promising node classification performance.

\section{Methodology}\label{sec:cal}
In this section, we firstly introduce notations, symbols, and necessary background about GNN models. We then present our idea of N2N mutual information maximization and its link to graph smoothing. Finally, we elaborate the proposed TAPS strategy.	

\subsection{GNN Framework}\label{sec:gfm}
We denote a graph as $\mathcal{G}=(\mathcal{V}, \mathcal{E}, \mathbf{A})$ with a set of nodes $\mathcal{V}$, a set of edges $\mathcal{E}$, and an adjacency matrix $\mathbf{A}$. Each node $v\in \mathcal{V}$ has a feature vector $\vec{\boldsymbol v}\in \mathcal{X}$ with node feature space $\mathcal{X}$ of dimensionality $D$. GNNs utilize a neighbourhood aggregation scheme to learn latent node representation $\vec{\boldsymbol h}^{(l)} \in \mathbb{R}^{D^{(l)}}$ in $l$-th layer for each node $v$, and a prediction function is applied to the node representations of final hidden layer to predict the class label $y_{v}$ of each node $v$. Based on such notations, a commonly adopted pipeline for supervised GNNs can be defined as:
\vspace{-0.2cm}
\begin{equation}\label{eq:pipeline}
\begin{split}
    \vec{\boldsymbol s}_{i}^{(l-1)} &= {\rm AGGREGATION}(\{\vec{\boldsymbol h}_{j}^{(l-1)}:v_{j}\in \mathcal{N}_{i}\}), \\
    \vec{\boldsymbol h}_{i}^{(l)} &= {\rm COMBINE}(\{\vec{\boldsymbol s}_{i}^{(l-1)}, \vec{\boldsymbol h}_{i}^{(l-1)}\}), \\
    \mathcal{L}_{{\rm CE}} &= {\rm PREDICTION}(\{\vec{\boldsymbol h}_{i}^{(L)}, y_{v_{i}}\}),
\end{split}
\end{equation}
where the AGGREGATION function can be any form of aggregators such as \emph{mean, max, sum, attention}, and \emph{ensemble} that learn the neighbourhood representation $\vec{\boldsymbol s}_{i}^{(l-1)}$ from the set $\{\vec{\boldsymbol h}_{j}^{(l-1)}:v_{j}\in \mathcal{N}_{i}\}$ based on the neighbourhood $\mathcal{N}_{i}$, and the COMBINE function updates $\vec{\boldsymbol h}_{i}^{(l-1)}$ to a new representation $\vec{\boldsymbol h}_{i}^{(l)}$ in $l$-th layer by combining $\vec{\boldsymbol s}_{i}^{(l-1)}$ with $\vec{\boldsymbol h}_{i}^{(l-1)}$. A $L$-layer GNN iterates the above two operations $L$ times and the PREDICTION function is applied in the output layer for node classification. A de-facto loss function for the PREDICTION layer is Cross-Entropy (CE) loss.

\subsection{Node-to-Neighbourhood (N2N) Mutual Information Maximization}\label{sec:n2n}
We learn topology-aware node representation by maximizing the mutual information between the hidden representations of nodes and their neighbourhood, which is partially motivated by knowledge distillation~\cite{park2019relational} in computer vision. In this section, we firstly present the definition of the N2N mutual information, which is followed by the optimization of the mutual information and its link to graph smoothing.

We denote the Probability Density Function (PDF) of the node representation $\vec{\boldsymbol h}_{i}^{(l)}$ over the feature space $\mathcal{X}^{D^{(l)}}$ in $[0,1]^{D^{(l)}}$ as $p(H(\bm{x})^{(l)})$ , where $\bm{x}\in \mathcal{X}^{D^{(l)}}$ and $H(\cdot)^{(l)}$ is a mapping function from $\bm{x}$ to $\vec{\boldsymbol h}_{i}^{(l)}$; the PDF of the neighbourhood representation $\vec{\boldsymbol s}_{i}^{(l)}$ as $p(S(\bm{x})^{(l)})$ with the mapping function $S(\cdot)^{(l)}$ from $\bm{x}$ to $\vec{\boldsymbol s}_{i}^{(l)}$; and the joint PDF between node and neighbourhood is $p(S(\bm{x})^{(l)}, H(\bm{x})^{(l)})$. We define the mutual information between the node representations and their corresponding neighbourhood representation as:
\vspace{-0.2cm}
\begin{equation}\label{eq:mutual-information}
\begin{split}
    &I(S(\bm{x})^{(l)};H(\bm{x})^{(l)})= \\
    &\underset{\mathcal{X}^{D^{(l)}}}{\int}p(S(\bm{x})^{(l)},H(\bm{x})^{(l)})\cdot\log\frac{p(S(\bm{x})^{(l)},H(\bm{x})^{(l)})}{p(S(\bm{x})^{(l)})\cdot p(H(\bm{x})^{(l)})}d\bm{x}.
\end{split}
\end{equation}
This operation encourages each node representation to distill the contextual information presented in its neighbourhood representation. However, mutual information is notoriously difficult to compute, particularly in continuous and high-dimensional space. Fortunately, scalable estimation enabling efficient computation of mutual information is made possible through Mutual Information Neural Estimation (MINE)~\cite{belghazi2018mutual}, which converts mutual information maximization into minimizing the InfoNCE loss~\cite{oord2018representation}. The surrogate InfoNCE loss function of the N2N mutual information in Eq.~(\ref{eq:mutual-information}) is defined as:
\vspace{-0.2cm}
\begin{equation}\label{eq:infonce}
\begin{split}
    &\mathcal{L}_{{\rm InfoNCE}}=\\
    &-\mathbb{E}_{v_{i}\in \mathcal{V}}\left[\log\frac{\exp({\rm sim}(\vec{\boldsymbol s}_{i}^{(l)}, \vec{\boldsymbol h}_{i}^{(l)})/\tau)}{\sum_{v_{k}\in \mathcal{V}}\exp({\rm sim}(\vec{\boldsymbol h}_{k}^{(l)}, \vec{\boldsymbol h}_{i}^{(l)})/\tau)}\right], \\
\end{split}
\end{equation}
which estimates the mutual information via node sampling, where the ${\rm sim}(\cdot, \cdot)$ function denotes the cosine similarity, the ${\rm exp}(\cdot)$ function implies the exponential function, and $\tau$ is the temperature parameter. The positive pair is $(\vec{\boldsymbol s}_{i}^{(l)}, \vec{\boldsymbol h}_{i}^{(l)})$ and the negative pair is $(\vec{\boldsymbol h}_{k}^{(l)}, \vec{\boldsymbol h}_{i}^{(l)})_{i\neq k}$.

In essence, maximizing $I(S(\bm{x})^{(l)};H(\bm{x})^{(l)})$ can play the role of graph smoothing, which has proven to be able to benefit node/graph prediction. To elaborately prove this point, we resort to a feature smoothness metric in~\cite{hou2019measuring}:
\vspace{-0.2cm}
\begin{equation}\label{eq:feature-smoothness}
    \delta_{f}^{(l)}=\frac{\|\sum_{v_{i}\in \mathcal{V}}(\sum_{v_{j}\in \mathcal{N}_{i}}(\vec{\boldsymbol h}_{i}^{(l)}-\vec{\boldsymbol h}_{j}^{(l)}))^{2}\|_{1}}{|\mathcal{E}|\cdot D^{(l)}},
\end{equation}
where $\|\cdot\|_{1}$ is the Manhattan norm. The work~\cite{hou2019measuring} further proposes that the information gain from the neighbourhood representation $\vec{\boldsymbol s}_{i}^{(l)}$ is defined as the Kullback-Leibler divergence:
\vspace{-0.2cm}
\begin{equation}\label{eq:kl-divergence}
\begin{split}
    &D_{KL}(S(\bm{x})^{(l)}\|H(\bm{x})^{(l)})= \\ &\underset{\mathcal{X}^{D^{(l)}}}{\int}p(S(\bm{x})^{(l)})\cdot\log\frac{p(S(\bm{x})^{(l)})}{p(H(\bm{x})^{(l)})}d\bm{x}, \\
\end{split}
\end{equation}
which is positively correlated to the feature smoothness metric $\delta_{f}^{(l)}$, \textit{i.e.}, $D_{KL}(S(\bm{x})^{(l)}\|H(\bm{x})^{(l)})\sim \delta_{f}^{(l)}$. This standpoint implies that a large feature smoothness value $\delta_{f}^{(l)}$ means significant disagreement between node representations $\{\vec{\boldsymbol h}_{i}^{(l)}\}$ and their corresponding neighbourhood representations $\{\vec{\boldsymbol s}_{i}^{(l)}\}$. This inspires the following theorem (Ref. Appendix~\ref{appe:theo:kl-mut} for proof):
\vspace{-0.2cm}
\begin{theorem}\label{theo:kl-mut}
    For a graph $\mathcal{G}$ with the set of features $\mathcal{X}^{D^{(l)}}$ in space $[0,1]^{D^{(l)}}$, the information gain represented by $D_{KL}(S(\bm{x})^{(l)}\|H(\bm{x})^{(l)})$ is negatively correlated to the mutual information $I(S(\bm{x})^{(l)};H(\bm{x})^{(l)})$ and thus maximizing $I(S(\bm{x})^{(l)};H(\bm{x})^{(l)})$ essentially minimizes $D_{KL}(S(\bm{x})^{(l)}\|H(\bm{x})^{(l)})$ and $\delta_{f}^{(l)}$, which attains the goal of graph smoothing:
    \vspace{-0.2cm}
    \begin{equation}
    \begin{split}
        I(S(\bm{x})^{(l)};H(\bm{x})^{(l)})&\sim \frac{1}{D_{KL}(S(\bm{x})^{(l)}\|H(\bm{x})^{(l)})} \\
        &\sim \frac{1}{\delta_{f}^{(l)}}. \\
    \end{split}
    \end{equation}
\end{theorem}

\subsection{Topology-Aware Positive Sampling (TAPS)}\label{sec:taps}
Up to now, we obtain the neighbourhood representation $\vec{\boldsymbol s}_{i}^{(l)}$ of a node by applying the AGGREGATION function to all neighbours of the node. This solution may suffer from two issues. Firstly, the whole neighbourhood can include redundant or even noisy information. Secondly, the aggregation operation is computationally expensive. To address these two problems, we propose a TAPS strategy for self-supervised node representation learning. The basic idea is that we measure the topological dependencies between a node and its neighbours and sample positives of the node based on the ranked dependency values.

For a node $v_i$, we use a variable $X_i$ to represent its topological information. $X_i$ can take the value of either $\mathcal{N}_i$ or $\overline{\mathcal{N}_i}=\mathcal{V}-\mathcal{N}_i$, where the former corresponds to the neighbourhood information and the latter is the contextual information complementary to the neighbourhood. Based on the definition of $X_i$, we define $p(X_{i}=\mathcal{N}_{i})=\frac{|\mathcal{N}_{i}|}{|\mathcal{V}|}$ and probability $p(X_{i}=\overline{\mathcal{N}_{i}})=\frac{|\mathcal{V}-\mathcal{N}_{i}|}{|\mathcal{V}|}$, where $|\cdot|$ is the cardinality function. Basically $p(X_{i}=\mathcal{N}_{i})$ indicates when we sample a node randomly on the graph, the probability that the node will fall into the neighbourhood of $v_{i}$. Furthermore, for two neighbouring nodes $v_i$ and $v_j$, we can define the following joint probabilities:
\vspace{-0.2cm}
\begin{equation}\label{eq:dist}
\begin{split}
    p(X_{i}=\mathcal{N}_{i}, X_{j}=\mathcal{N}_{j})&=\frac{|\mathcal{N}_{i}\cap \mathcal{N}_{j}|}{|\mathcal{V}|}, \\
    p(X_{i}=\mathcal{N}_{i}, X_{j}=\overline{\mathcal{N}_{j}})&=\frac{|\mathcal{N}_{i}\cap (\mathcal{V}-\mathcal{N}_{j})|}{|\mathcal{V}|}, \\
    p(X_{i}=\overline{\mathcal{N}_{i}}, X_{j}=\mathcal{N}_{j})&=\frac{|(\mathcal{V}-\mathcal{N}_{i})\cap \mathcal{N}_{j}|}{|\mathcal{V}|}, \\
    p(X_{i}=\overline{\mathcal{N}_{i}}, X_{j}=\overline{\mathcal{N}_{j}})&=\frac{|(\mathcal{V}-\mathcal{N}_{i})\cap (\mathcal{V}-\mathcal{N}_{j})|}{|\mathcal{V}|},
\end{split}
\end{equation}
where $p(X_{i}=\mathcal{N}_{i}, X_{j}=\mathcal{N}_{j})$ is the probability that the randomly selected node will fall into the intersected neighbours of $v_{i}$ and $v_{j}$. 
Motivated by mutual information, we define the graph-structural dependency between $v_i$ and $v_j$ as:
\vspace{-0.2cm}
\begin{definition}\label{definition:gsd}
    Graph-structural dependency between neighbouring node $v_{i}$ and $v_{j}$ is defined as:
    \vspace{-0.2cm}
    \begin{equation}\label{eq:tami}
    \begin{split}
        I(X_{i};X_{j})=&\sum_{X_{i}} \sum_{X_{j}} p(X_{i}, X_{j})\cdot\log\frac{p(X_{i}, X_{j})}{p(X_{i})\cdot p(X_{j})}, \\
        &s.t.\;v_{j}\in \mathcal{N}_{i}. \\
    \end{split}
    \end{equation}
\end{definition}
The graph-structural dependency value above basically measures the topological similarity of two nodes. A large value suggests the strong dependency between two nodes.

In TAPS strategy, we select positives of $v_{i}$ by ranking the dependency values between the neighbouring nodes and $v_i$ and then obtain the neighbourhood representation $\vec{\boldsymbol s}_{i}^{(l)}$ by applying aggregator, \textit{e.g.}, \emph{mean}, to the selected positives. In particular, when only one positive is selected, we directly select the node $v_j$ with maximum dependency value to $v_i$ and thus avoid the expensive aggregation operation. Meanwhile, because the topology structure of a graph relies on the adjacency matrix only, TAPS allows us to perform positive sampling upfront, which can avoid the positive sampling overhead during training.

\section{Training Frameworks}\label{sec:tf}
According to the relationship among GNN encoders, \emph{pretext} tasks, and \emph{downstream} tasks, there are three types of graph-based self-supervised training schemes~\cite{liu2021graph}. The first type is Pre-training and Fine-tuning (PT\&FT). The pre-training stage firstly initializes the parameters of the GNN encoder with \emph{pretext} tasks. After this, this pre-trained GNN encoder is fine-tuned under the supervision of specific \emph{downstream} tasks. The second is Joint Learning (JL) scheme, where the GNN encoder, \emph{pretext} and \emph{downstream} tasks are trained jointly. The last type is Unsupervised Representation Learning (URL). Akin to PT\&FT, URL also follows a two-stage training scheme, where the first stage trains the GNN encoder based on the \emph{pretext} task but in the second \emph{downstream} task stage, the GNN encoder is frozen. In our work, we adopt both JL and URL pipelines to train and evaluate our network.
\begin{figure*}[!t]
    \centering
    \includegraphics[scale=0.4]{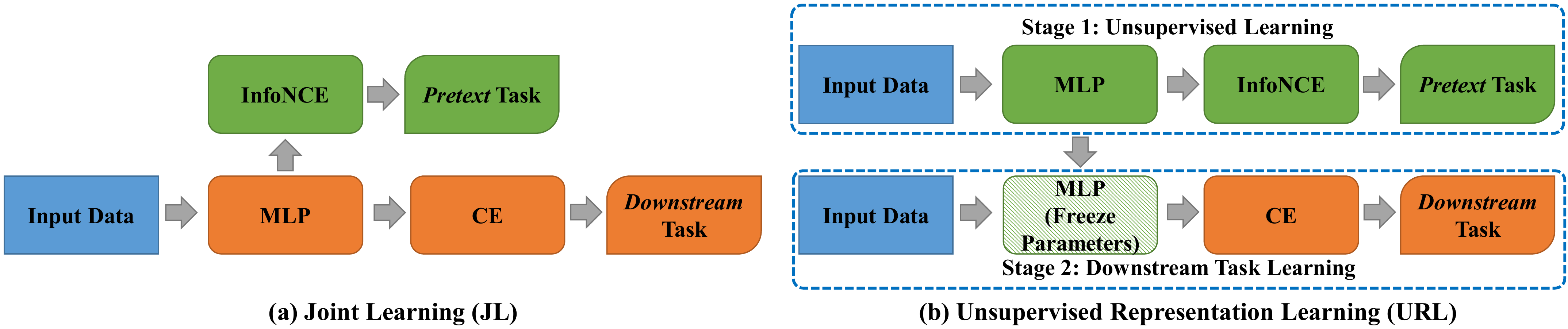}
    \vspace{-0.5cm}
    \caption{Illustration of two training pipelines adopted for the proposed model.}\label{fig:jl-url}
    \vspace{-0.5cm}
\end{figure*}

\textbf{JL Training Framework.}~Fig.~\ref{fig:jl-url}~(a) illustrates our JL training pipeline. As can be seen, unlike most existing graph-based SSL work that uses GNN as node/graph encoder, we simply use a shallow MLP as encoder, which is more efficient. In JL scheme, we apply InfoNCE loss and Cross-Entropy loss jointly on top of the node representations obtained as output of the MLP encoder:
\vspace{-0.2cm}
\begin{align}
\mathcal{L}=(1-\alpha)\mathcal{L}_{{\rm CE}} + \alpha\mathcal{L}_{{\rm InfoNCE}},
\end{align}
where $\alpha$ is a trade-off parameter used to balance the two loss functions.

\textbf{URL Training Framework.}~The URL framework, as shown in Fig.~\ref{fig:jl-url}~(b), involves two training stages: the pre-training \emph{pretext} task trains the MLP encoder using InfoNCE loss $\mathcal{L}_{{\rm InfoNCE}}$, and the \emph{downstream} task learns linear node classifiers using Cross-Entropy loss $\mathcal{L}_{{\rm CE}}$.

\section{Experiments}\label{sec:exper}
In this section, we start by introducing the experimental setups including datasets, comparing methods, and the implementation details. Then we present the performance comparison to existing methods. Finally, ablation studies are presented to reveal other appealing properties of the proposed methods.

\subsection{Experimental Setup}
\textbf{Datasets.}~We conduct experiments on six real-world node classification datasets: Cora~\cite{you2020layer}, Pubmed~\cite{you2020layer}, Citeseer~\cite{you2020layer}, Amazon Photo~\cite{shchur2018pitfalls}, Coauthor CS~\cite{shchur2018pitfalls}, and Coauthor Physics~\cite{shchur2018pitfalls}. The first three datasets are constructed as citation networks\footnote{For these three datasets, we use the train/val/test splits in ~\cite{hou2019measuring,you2020layer}.}, Amazon Photo is for the Amazon co-purchase graph, and Coauthor CS \& Physics are for co-authorship graphs. All of the above six graphs are connected and undirected.

\textbf{Our Models.}~We denote N2N-TAPS-$x$ as our model sampling top-$x$ positive neighbours based on TAPS, \textit{e.g.}, N2N-TAPS-1 samples the one neighbour with maximum dependency value as positive. We evaluate our methods with \numrange{1}{5} positives to inspect how the positive size influences the node classification performance. We use N2N-random-$1$ as our baseline where one positive is sampled randomly from the neighbourhood of a node. By default, we aggregate all the neighbours of a node using \emph{mean} aggregator as positive, which is denoted as N2N.

\textbf{Existing Methods.}~We employ three types of methods for comparison: GNN methods, conventional SSL methods, and GCL methods. For each type, some representatives are chosen. The GNN solutions include GCN~\cite{kipf2017semi}, SGC~\cite{wu2019simplifying}, L2-GCN~\cite{you2020layer}, GraphSAGE-Mean~\cite{hamilton2017inductive}, FastGCN~\cite{chen2018fastgcn}, GAT~\cite{velickovic2019graph}, SplineCNN~\cite{fey2018splinecnn}, PNA~\cite{corso2020principal}, and CS-GCN~\cite{hou2019measuring}. For conventional SSL methods, we incorporate DeepWalk~\cite{perozzi2014deepwalk} and node2vec~\cite{grover2016node2vec} for comparison. The state-of-the-art GCL methods used for comparison include DGI~\cite{velivckovic2018deep}, GMI~\cite{peng2020graph}, MVGRL~\cite{hassani2020contrastive}, InfoGraph~\cite{sun2020infograph}, and Graph-MLP~\cite{hu2021graph}. The key properties of such SSL/GCL methods and our methods are summarized in Table~\ref{table:differences} from three perspectives: learning type, training framework, and encoder type.
\begin{table}[!h]
    \vspace{-0.2cm}
    \caption{Summaries of typical SSL/GCL methods and our method.}\label{table:differences}
    \vspace{-0.2cm}
    \centering
    \resizebox{84mm}{!}{
    \begin{tabular}{l|ccc}
    \toprule[1.2pt]
    \textbf{Model} & \textbf{Learning Type} & \textbf{Framework} & \textbf{Encoder} \\
    \hline
    DeepWalk~\cite{perozzi2014deepwalk} & \multirow{2}{*}{SSL} & URL & Shallow MLP \\
    node2vec~\cite{grover2016node2vec} &  & URL & Shallow MLP \\
    \hline
    DGI~\cite{velivckovic2018deep} & \multirow{5}{*}{GCL} & URL & GCN \\
    GMI~\cite{peng2020graph} &  & URL & GCN \\
    MVGRL~\cite{hassani2020contrastive} &  & URL & GCN \\
    InfoGraph~\cite{sun2020infograph} &  & URL & GIN \\
    Graph-MLP~\cite{hu2021graph} &  & JL & Shallow MLP \\
    \hline
    N2N-TAPS-$x$ & GCL & JL \& URL & Shallow MLP \\
    \bottomrule[1.2pt]
    \end{tabular}
    }
    \vspace{-0.2cm}
\end{table}

\textbf{Implementation Details.}~For fair comparison, we follow the common practice to fix the number of hidden layers in our method and the compared GNNs and GCL encoders to be 2. For all datasets, we set the dimensionality of the hidden layer to be $512$. Some other important hyper-parameters include: dropout ratio is $0.6$ for Cora, Citeseer, and Coauthor CS, $0.2$ for Pubmed, $0.4$ for Amazon Photo, and $0.5$ for Coauthor Physics; L2-regularization is $0.01$ for Cora, Citeseer, and Coauthor CS, $0.001$ for Pubmed and Amazon Photo, and $0.05$ for Coauthor Physics; training epochs are $1000$ for all datasets; learning rate is $0.01$ for Pubmed and $0.001$ for the other datasets; nonlinear activation is ${\rm ReLU}$ function. For N2N-TAPS-$x$ (JL), $\alpha$ is set to $0.9$ for Cora, Pubmed, and Citeseer, $0.99$ for Amazon Photo and Coauthor CS, and $0.999$ for Coauthor Physics; temperature $\tau$ is $5$ for Cora, Pubmed, Citeseer, and Amazon Photo, $100$ for Coauthor CS, and $30$ for Coauthor Physics. The temperature $\tau$ of N2N-TAPS-$x$ (URL) is $5$ for all datasets. These hyper-parameters are determined via cross-validation. We implement our models by Tensorflow 2.4. All of the experiments are performed on a machine with Intel 8-core i7-10870H CPU (2.20GHz), 32GB CPU memory, and one GeForce RTX 3080 Laptop card (16GB GPU memory). For each dataset, we run all models five times, and the mean and standard deviation of the micro-f1 score are used as the evaluation metric.

\subsection{Overall Results}
\begin{table*}[!t]
    \caption{Performance comparison to existing methods on six datasets. The results for the comparing methods are obtained either by running the publicly released code or through our own implementation. Mean and standard deviation of 5-fold Micro-f1 scores are reported as evaluation metric. The best results for each dataset are highlighted in \textbf{bold}.}\label{table:overall}
    \vspace{-0.2cm}
    \centering
    \resizebox{120mm}{!}{
    \begin{tabular}{l|cccccc}
    \toprule[1.2pt]
    \multirow{2}{*}{\textbf{Model}} & \multirow{2}{*}{\textbf{Cora}} & \multirow{2}{*}{\textbf{Pubmed}} & \multirow{2}{*}{\textbf{Citeseer}} & \textbf{Amazon} & \textbf{Coauthor} & \textbf{Coauthor} \\
     & & & & \textbf{Photo} & \textbf{CS} & \textbf{Physics} \\
    \hline
    GCN~\cite{kipf2017semi} & 84.72$\pm$0.08 & 87.02$\pm$0.06 & 78.20$\pm$0.15 & 90.02$\pm$0.10 & 90.52$\pm$0.21 & 91.04$\pm$0.06 \\
    SGC~\cite{wu2019simplifying} & 84.25$\pm$0.10 & 86.68$\pm$0.06 & 77.65$\pm$0.12 & 89.36$\pm$0.16 & 90.03$\pm$0.15 & 90.12$\pm$0.08 \\
    L2-GCN~\cite{you2020layer} & 84.56$\pm$0.03 & 86.80$\pm$0.06 & 77.06$\pm$0.08 & 89.16$\pm$0.26 & 90.52$\pm$0.04 & 91.15$\pm$0.10 \\
    GraphSAGE-Mean~\cite{hamilton2017inductive} & 85.04$\pm$0.12 & 87.15$\pm$0.14 & 77.82$\pm$0.15 & 90.05$\pm$0.04 & 90.40$\pm$0.08 & 90.89$\pm$0.12 \\
    FastGCN~\cite{chen2018fastgcn} & 84.08$\pm$0.04 & 86.92$\pm$0.08 & 77.65$\pm$0.05 & 88.65$\pm$0.12 & 90.00$\pm$0.05 & 89.60$\pm$0.25 \\
    GAT~\cite{velickovic2019graph} & 85.23$\pm$0.15 & 87.85$\pm$0.14 & 78.05$\pm$0.26 & 86.78$\pm$0.26 & 91.10$\pm$0.10 & 91.17$\pm$0.15 \\
    SplineCNN~\cite{fey2018splinecnn} & 85.45$\pm$0.16 & 87.82$\pm$0.08 & 78.83$\pm$0.13 & 89.08$\pm$0.18 & 91.13$\pm$0.20 & 90.82$\pm$0.16 \\
    PNA~\cite{corso2020principal} & 85.40$\pm$0.12 & 87.20$\pm$0.07 & 78.28$\pm$0.05 & 90.23$\pm$0.14 & 91.35$\pm$0.16 & 90.68$\pm$0.14 \\
    CS-GCN~\cite{hou2019measuring} & 85.14$\pm$0.04 & 87.75$\pm$0.08 & 78.85$\pm$0.16 & 90.12$\pm$0.14 & 91.14$\pm$0.08 & 90.23$\pm$0.08 \\
    DeepWalk~\cite{perozzi2014deepwalk} & 77.84$\pm$0.12 & 86.52$\pm$0.12 & 60.24$\pm$0.28 & 83.95$\pm$0.28 & 84.75$\pm$0.28 & 86.25$\pm$0.22 \\
    node2vec~\cite{grover2016node2vec} & 75.15$\pm$0.06 & 85.20$\pm$0.02 & 65.52$\pm$0.18 & 84.16$\pm$0.18 & 85.28$\pm$0.18 & 85.58$\pm$0.18 \\
    DGI~\cite{velivckovic2018deep} & 85.08$\pm$0.05 & 87.03$\pm$0.08 & 78.82$\pm$0.15 & 90.06$\pm$0.17 & 90.85$\pm$0.08 & 89.88$\pm$0.20 \\
    GMI~\cite{peng2020graph} & 85.26$\pm$0.08 & 87.26$\pm$0.16 & 78.69$\pm$0.16 & 89.25$\pm$0.10 & 90.80$\pm$0.20 & 90.05$\pm$0.08 \\
    MVGRL~\cite{hassani2020contrastive} & 85.38$\pm$0.06 & 87.25$\pm$0.12 & 78.08$\pm$0.06 & 88.23$\pm$0.10 & 90.62$\pm$0.08 & 89.68$\pm$0.18 \\
    InfoGraph~\cite{sun2020infograph} & 84.32$\pm$0.08 & 87.56$\pm$0.12 & 78.85$\pm$0.12 & 90.10$\pm$0.20 & 90.42$\pm$0.08 & 90.18$\pm$0.12 \\
    Graph-MLP~\cite{hu2021graph} & 82.50$\pm$0.10 & 87.25$\pm$0.13 & 78.86$\pm$0.08 & 89.25$\pm$0.14 & 90.25$\pm$0.20 & 89.45$\pm$0.10 \\
    \hline
    N2N-Random-1 (JL) & 83.46$\pm$0.18 & 86.20$\pm$0.08 & 76.85$\pm$0.30 & 86.25$\pm$0.15 & 89.65$\pm$0.06 & 88.75$\pm$0.12 \\
    N2N-TAPS-1 (JL) & 85.46$\pm$0.08 & \textbf{91.08$\pm$0.16} & 80.24$\pm$0.13 & 89.90$\pm$0.08 & 92.07$\pm$0.06 & 90.70$\pm$0.20 \\
    N2N-TAPS-2 (JL) & 86.36$\pm$0.16 & 90.76$\pm$0.14 & 80.34$\pm$0.06 & 89.77$\pm$0.07 & 92.11$\pm$0.08 & 90.28$\pm$0.24 \\
    N2N-TAPS-3 (JL) & 86.74$\pm$0.14 & 90.74$\pm$0.05 & 80.64$\pm$0.14 & 89.71$\pm$0.10 & 92.21$\pm$0.07 & 90.81$\pm$0.17 \\
    N2N-TAPS-4 (JL) & 86.52$\pm$0.15 & 90.78$\pm$0.07 & \textbf{81.06$\pm$0.11} & 89.89$\pm$0.08 & 92.27$\pm$0.07 & 90.42$\pm$0.22 \\
    N2N-TAPS-5 (JL) & 87.10$\pm$0.08 & 90.78$\pm$0.20 & 80.84$\pm$0.11 & 89.99$\pm$0.06 & 92.37$\pm$0.05 & 91.38$\pm$0.12 \\
    N2N (JL) & 87.52$\pm$0.20 & 90.92$\pm$0.09 & 80.90$\pm$0.21 & 90.12$\pm$0.26 & \textbf{93.06$\pm$0.07} & 91.41$\pm$0.09 \\
    \hline
    N2N-Random-1 (URL) & 82.50$\pm$0.10 & 84.08$\pm$0.12 & 75.06$\pm$0.12 & 86.07$\pm$0.12 & 88.23$\pm$0.08 & 86.56$\pm$0.04 \\
    N2N-TAPS-1 (URL) & 84.66$\pm$0.11 & 88.42$\pm$0.07 & 77.44$\pm$0.07 & 89.40$\pm$0.05 & 91.72$\pm$0.03 & 90.35$\pm$0.06 \\
    N2N-TAPS-2 (URL) & 85.60$\pm$0.08 & 89.26$\pm$0.08 & 78.56$\pm$0.08 & 90.49$\pm$0.04 & 91.56$\pm$0.03 & 90.68$\pm$0.02 \\
    N2N-TAPS-3 (URL) & 87.96$\pm$0.08 & 89.24$\pm$0.05 & 78.36$\pm$0.04 & 90.61$\pm$0.05 & 91.53$\pm$0.08 & 91.20$\pm$0.08 \\
    N2N-TAPS-4 (URL) & 88.04$\pm$0.11 & 89.32$\pm$0.09 & 78.54$\pm$0.07 & 90.35$\pm$0.07 & 92.03$\pm$0.06 & 90.89$\pm$0.05 \\
    N2N-TAPS-5 (URL) & 87.84$\pm$0.09 & 89.88$\pm$0.05 & 79.08$\pm$0.07 & 90.65$\pm$0.07 & 91.99$\pm$0.04 & 91.52$\pm$0.02 \\
    N2N (URL) & \textbf{88.20$\pm$0.05} & 89.30$\pm$0.04 & 79.54$\pm$0.02 & \textbf{91.08$\pm$0.11} & 92.31$\pm$0.05 & \textbf{91.77$\pm$0.08} \\
    \bottomrule[1.2pt]
    \end{tabular}
    }
    \vspace{-0.2cm}
\end{table*}
Table~\ref{table:overall} shows the performance comparison between our method and other selected methods. From the results we have the following observations: (1). Our N2N models, either N2N(JL) or N2N(URL), consistently outperform the comparing methods on all the six datasets. The margins can go up to $3\%$ on datasets like Cora, Pubmed, and Coauthor CS. This shows the competitiveness of our N2N mutual information maximization strategy over GNNs and other GCL based solutions for node representation learning. Another issue worth mentioning is that since our N2N based methods avoid topological augmentations and simply use MLP as node encoders, our methods are more efficient in terms of training and inference. (2). Within the N2N family, we generally observe improvement when sampling more positives based on TAPS but the improvements are marginal. This demonstrates the potential of N2N-TAPS-1 as it avoids neighbourhood aggregation operation which is known to be expensive. However, when the single positive is sampled randomly from the neighbourhood, the performance drops significantly. This result shows the proposed TAPS strategy indeed can sample topologically meaningful positives. (3). Within the existing methods, the GCL solutions have comparable performance or even slightly better performance comparing to the fully supervised GNN variants. This observation shows SSL can be a promising alternative in graph-based representation learning.

\subsection{Ablation Studies}
In this section, we conduct additional ablation studies to reveal other appealing properties of the proposed methods.

\begin{table}[!h]
    \vspace{-0.2cm}
    \caption{Performance of N2N (JL) model with random positive sampling on Amazon Photo and Coauthor Physics. We vary the positive sampling size from 1 to 5 and for each sampling size we run the experiments three times with different random seeds.}\label{table:random}
    \vspace{-0.2cm}
    \centering
    \resizebox{84mm}{!}{
    \begin{tabular}{l|ccc|ccc}
    \toprule[1.2pt]
    \textbf{Model} & \multicolumn{3}{c}{\textbf{Amazon Photo}} & \multicolumn{3}{|c}{\textbf{Coauthor Physics}} \\
    \hline
    Random Seeds & 1 & 2 & 3 & 1 & 2 & 3 \\
    \hline
    N2N-Random-1 (JL) & 86.25$\pm$0.15 & 85.52$\pm$0.14 & 84.08$\pm$0.20 & 88.75$\pm$0.12 & 87.25$\pm$0.10 & 86.20$\pm$0.28 \\
    N2N-Random-2 (JL) & 87.68$\pm$0.28 & 85.74$\pm$0.18 & 85.62$\pm$0.38 & 88.62$\pm$0.24 & 87.28$\pm$0.14 & 86.02$\pm$0.10 \\
    N2N-Random-3 (JL) & 87.95$\pm$0.10 & 86.28$\pm$0.08 & 85.20$\pm$0.16 & 89.24$\pm$0.16 & 87.64$\pm$0.10 & 86.52$\pm$0.14 \\
    N2N-Random-4 (JL) & 88.25$\pm$0.26 & 86.06$\pm$0.20 & 86.21$\pm$0.42 & 88.82$\pm$0.10 & 86.52$\pm$0.20 & 87.30$\pm$0.10 \\
    N2N-Random-5 (JL) & 88.30$\pm$0.12 & 85.52$\pm$0.14 & 86.56$\pm$0.30 & 89.65$\pm$0.20 & 87.82$\pm$0.15 & 87.22$\pm$0.14 \\
    \bottomrule[1.2pt]
    \end{tabular}
    }
    \vspace{-0.3cm}
\end{table}
\textbf{N2N (JL) based on Random Positive Sampling}.~To further justify the necessity and advantage of our TAPS strategy, we run experiments on random positive sampling by varying the sampling size from 1 to 5. We choose two datasets, \textit{i.e.}, Amazon Photo and Coauthor Physics, for this experiment because their average node degree $> 5$. For each sampling size, we run the experiments three times with different random seeds. The results are shown in Table~\ref{table:random}. From the table we can clearly observe that random positive sampling results in large performance variance which means random sampling fails to identify consistent and informative neighbours.




\begin{table}[!b]
    \vspace{-0.3cm}
    \caption{The time cost comparison among typical GNN/GCL methods and our N2N-TAPS-$x$ models. The number indicates the time consumption for training a method on a dataset for one epoch. The symbol $\dagger$ behind N2N-TAPS-$x$ models indicates inference time. The unit of time is milliseconds (ms).}\label{table:time-consumption}
    \vspace{-0.2cm}
    \centering
    \resizebox{84mm}{!}{
    \begin{tabular}{l|cccccc}
    \toprule[1.2pt]
    \multirow{2}{*}{\textbf{Model}} & \multirow{2}{*}{\textbf{Cora}} & \multirow{2}{*}{\textbf{Pubmed}} & \multirow{2}{*}{\textbf{Citeseer}} & \textbf{Amazon} & \textbf{Coauthor} & \textbf{Coauthor} \\
    & & & & \textbf{Photo} & \textbf{CS} & \textbf{Physics} \\
    \hline
    GCN~\cite{kipf2017semi} & 1.68 & 5.88 & 6.64 & 6.08 & 10.06 & 11.31 \\
    GraphSAGE-Mean~\cite{hamilton2017inductive} & 1.56 & 6.52 & 6.46 & 6.28 & 9.65 & 12.06 \\
    FastGCN~\cite{chen2018fastgcn} & 1.21 & 5.73 & 6.89 & 5.24 & 9.06 & 10.04 \\
    DGI~\cite{velivckovic2018deep} & 1.82 & 7.20 & 8.56 & 8.25 & 12.00 & 15.68 \\
    GMI~\cite{peng2020graph} & 2.08 & 7.93 & 9.62 & 10.18 & 15.54 & 21.16 \\
    MVGRL~\cite{hassani2020contrastive} & 1.95 & 7.64 & 9.06 & 9.30 & 13.11 & 16.92 \\
    InfoGraph~\cite{sun2020infograph} & 1.86 & 8.65 & 11.16 & 11.42 & 12.82 & 15.20 \\
    Graph-MLP~\cite{hu2021graph} & 0.45 & 1.08 & 0.56 & 0.50 & 1.82 & 3.64 \\
    \hline
    N2N-TAPS-1 (JL) & 0.09 & 0.12 & 0.11 & 0.12 & 0.22 & 0.24 \\
    N2N-TAPS-5 (JL) & 0.15 & 0.14 & 0.18 & 0.19 & 0.46 & 0.63 \\
    N2N (JL)        & 0.30 & 0.19 & 0.19 & 0.20 & 1.19 & 1.68 \\
    \hline
    N2N-TAPS-1 (JL)$\dagger$ & 0.05 & 0.07 & 0.13 & 0.08 & 0.06 & 0.07 \\
    N2N-TAPS-5 (JL)$\dagger$ & 0.06 & 0.09 & 0.16 & 0.09 & 0.08 & 0.10 \\
    N2N (JL)$\dagger$        & 0.08 & 0.11 & 0.19 & 0.09 & 0.12 & 0.13 \\
    \hline
    N2N-TAPS-1 (URL) & 0.10 & 0.15 & 0.13 & 0.16 & 0.30 & 0.28 \\
    N2N-TAPS-5 (URL) & 0.18 & 0.18 & 0.16 & 0.25 & 0.42 & 0.54 \\
    N2N (URL)        & 0.22 & 0.25 & 0.18 & 0.38 & 1.25 & 1.80 \\
    \hline
    N2N-TAPS-1 (URL)$\dagger$ & 0.06 & 0.04 & 0.15 & 0.09 & 0.08 & 0.10 \\
    N2N-TAPS-5 (URL)$\dagger$ & 0.05 & 0.06 & 0.16 & 0.10 & 0.10 & 0.13 \\
    N2N (URL)$\dagger$        & 0.06 & 0.06 & 0.16 & 0.14 & 0.13 & 0.15 \\
    \bottomrule[1.2pt]
    \end{tabular}
    }
    \vspace{-0.2cm}
\end{table}
\textbf{Time Consumption.}~Our methods are expected to be more efficient comparing to existing work. On one hand, our work adopts MLP as node encoder and thus avoids the expensive node aggregation in the encoding phase. On the other hand, TAPS enables us to sample limited high-quality positives upfront. Especially, when one positive is selected, we fully get rid of the aggregation operation.

Table~\ref{table:time-consumption} shows the time consumption comparison. From the results we can see our methods can be orders of faster than the typical GNN and GCL based methods. Graph-MLP~\cite{hu2021graph} also adopts MLP as encoder but it aligns a node to all the nodes that can be reached from this node. This explains its slowness on large datasets such as CS and Physics.

\begin{figure}[!b]
    \centering
    \includegraphics[scale=0.3]{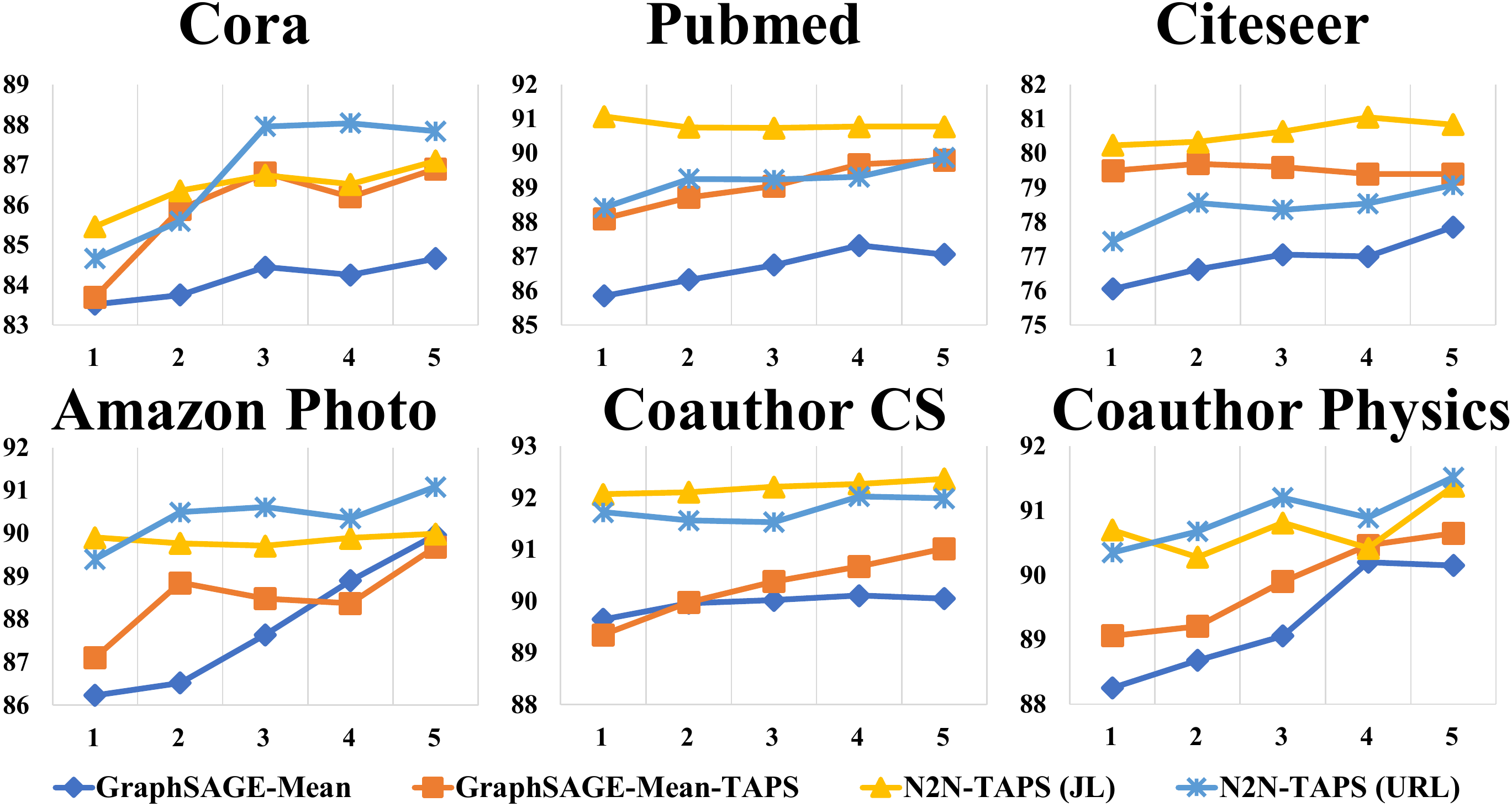}
    \vspace{-0.3cm}
    \caption{The performance comparison among GraphSAGE-Mean, GraphSAGE-Mean-TAPS, and N2N-TAPS-$x$ (JL and URL) based on varying number of sampled neighbours. Random neighbour sampling is used for GraphSAGE-Mean and TAPS is used for GraphSAGE-Mean-TAPS.}\label{fig:sampling}
    \vspace{-0.5cm}
\end{figure}
\textbf{Evaluation of TAPS Strategy.}~TAPS is an important component in our framework to ensure the quality and efficiency of positive sampling. In Table~\ref{table:overall} we have shown the advantage of TAPS over random sampling on our N2N-TAPS-1 model. In this section, we apply TAPS sampling to another sampling based GNN baseline, GraphSAGE-Mean, to verify if TAPS can be used as a general neighbourhood sampling strategy to identify informative neighbours. Fig.~\ref{fig:sampling} shows the results. By default, GraphSAGE-Mean~\cite{hamilton2017inductive} uses random sampling to select neighbours for aggregation, which has the risk of absorbing noisy information. We replace random sampling in GraphSAGE-Mean with TAPS and leave all the other implementation intact. Its performance is obviously boosted and generally using more neighbours can benefit the performance more. This observation shows us again that it is important to consider the structural dependencies to select useful neighbours to enrich node representations.

\begin{figure}[!t]
    \centering
    \includegraphics[scale=0.27]{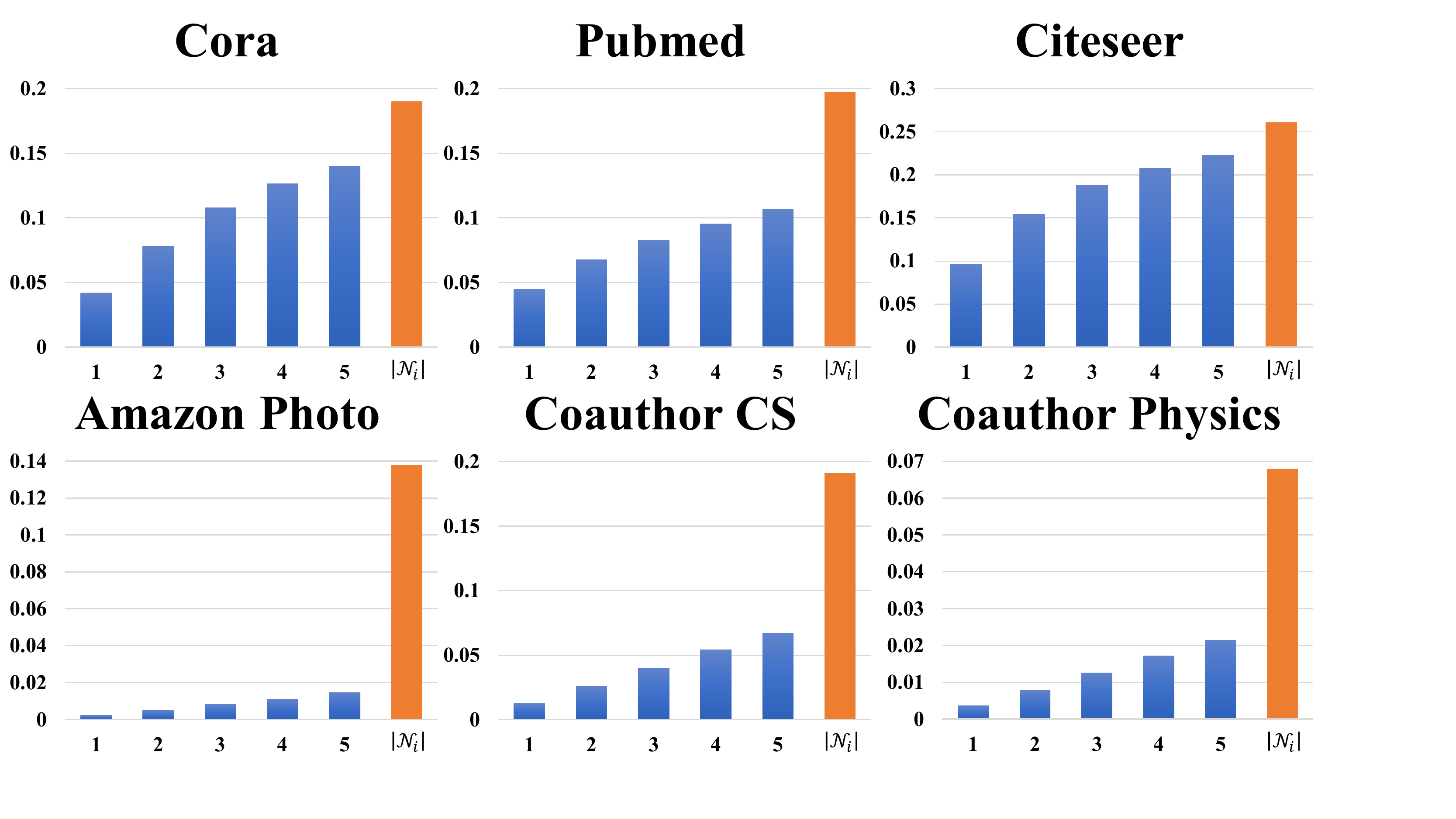}
    \vspace{-0.3cm}
    \caption{Label smoothness values obtained by using TAPS strategy to sample 1 to 5 neighbours (blue bars) and the label smoothness value obtained by considering all neighbours without any sampling strategy (orange bar).}\label{fig:label-smoothness}
    \vspace{-0.3cm}
\end{figure}
\begin{figure}[!t]
    \centering
    \includegraphics[scale=0.47]{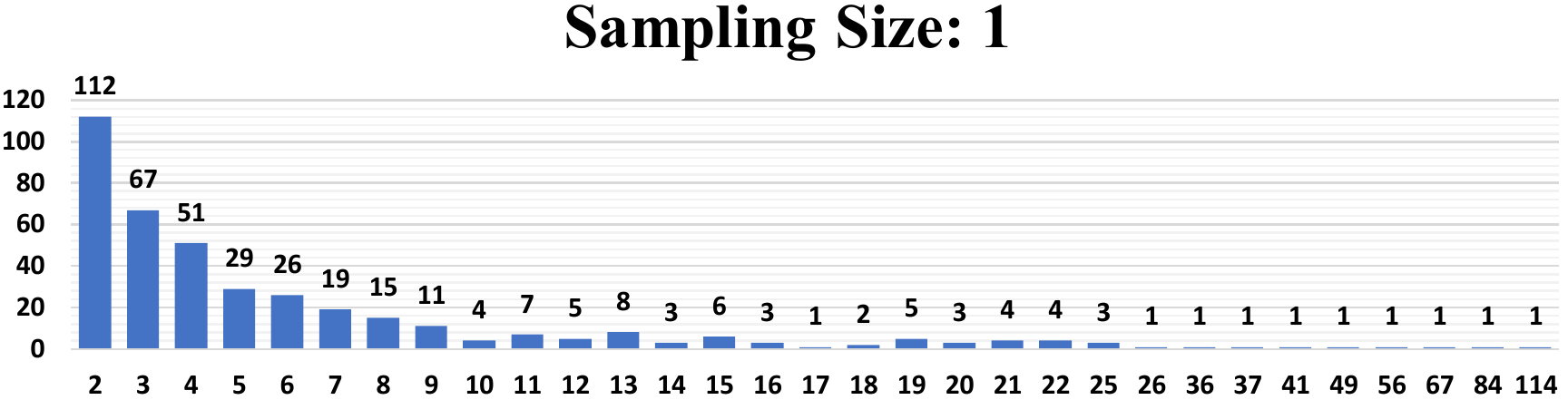}
    \vspace{-0.3cm}
    \caption{Subgraph statistics on Cora by using TAPS strategy to perform the subgraph partition. The horizontal axis indicates the number of nodes in the subgraph; the vertical axis implies the number of such subgraphs.}\label{fig:subgraph}
    \vspace{-0.5cm}
\end{figure}
\begin{figure}[!h]
    \centering
    \includegraphics[scale=0.21]{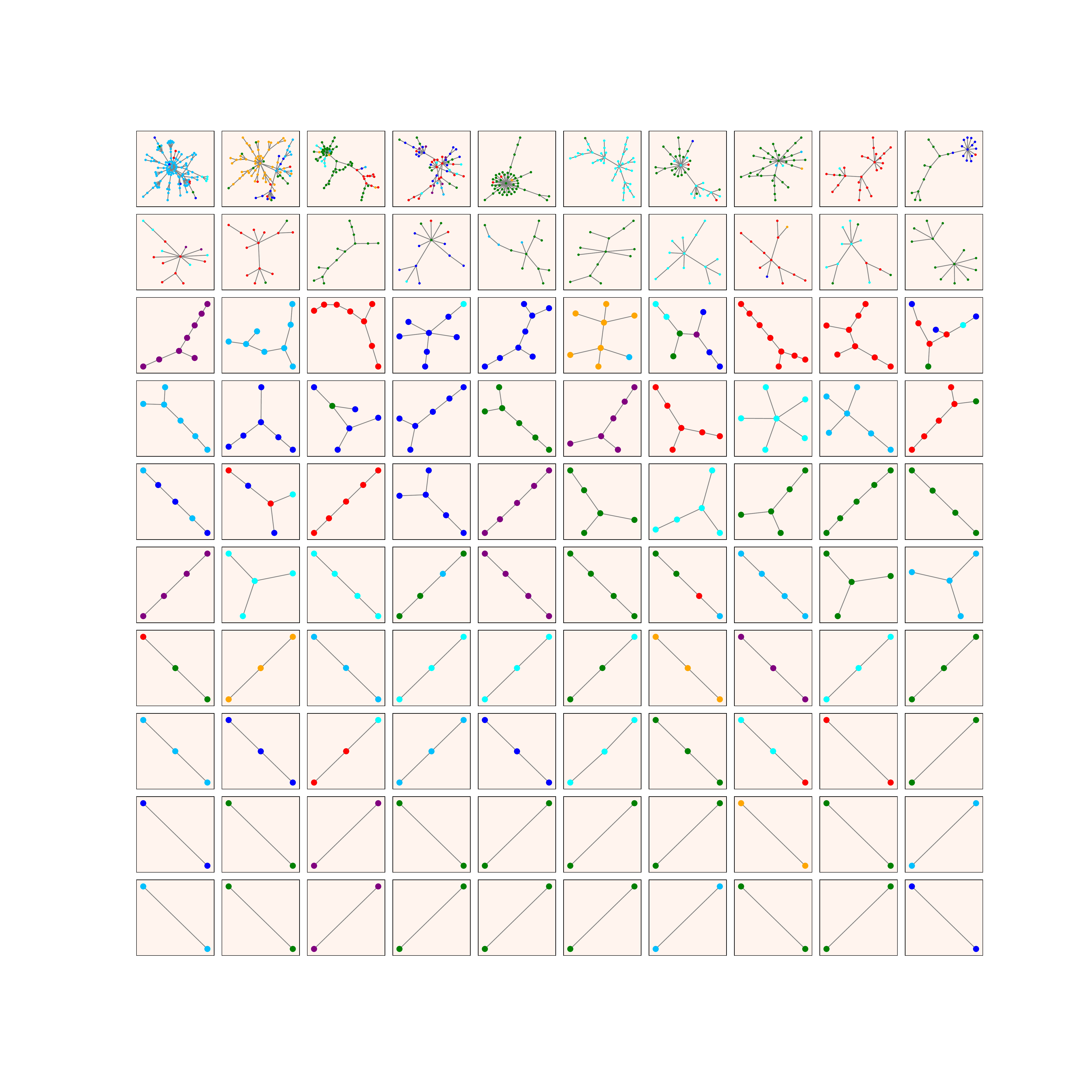}
    \vspace{-0.5cm}
    \caption{Visualization of part of the subgraphs derived from TAPS on Cora. We rank the sizes of the subgraphs and divide them into 10 intervals. For each row, we visualize the top-10 subgraphs within the corresponding interval from top (top $10\%$ interval) to bottom  ($80\% \sim 90\%$ interval). Different node colors represent different labels.}\label{fig:subgraph-shown}
    \vspace{-0.6cm}
\end{figure}
\textbf{Label Smoothness Analysis.}~To verify the quality of the neighbourhood sampling by using TAPS strategy, we introduce the label smoothness metric proposed in CS-GNN~\cite{hou2019measuring}: $\delta_{l}=\sum_{(v_{i},v_{j})\in \mathcal{E}}(1-\mathbb{I}(v_{i}\simeq v_{j}))/|\mathcal{E}|$, where $\mathbb{I}(\cdot)$ is an indicator function that has $\mathbb{I}(v_{i}\simeq v_{j})=1$ if $y_{v_{i}}=y_{v_{j}}$. Otherwise, $\mathbb{I}(v_{i}\simeq v_{j})=0$ when $y_{v_{i}}\neq y_{v_{j}}$. By virtue of the label smoothness, a large $\delta_{l}$ suggests that the nodes with different labels are regarded to be connected neighbours, while a small $\delta_{l}$ indicates a graph $\mathcal{G}$ possessing the higher-quality neighbourhood structure, \textit{i.e.}, most neighbours of a node have the same label as the node. Such high-quality neighbourhood can contribute homogeneous information gain to their corresponding central nodes~\cite{hou2019measuring}.

Fig.~\ref{fig:label-smoothness} shows that the label smoothness values gradually increases by expanding the sampling size from 1 to 5 with our TAPS strategy. Without any sampling strategy, the label smoothness value of the whole graph is the highest. This phenomenon suggests that our TAPS strategy can promote the neighbourhood sampling quality as the sampling size decreases, explaining why the proposed N2N-TAPS-1 model has competitive performance on some datasets.

TAPS strategy is essentially a subgraph partition scheme. A good partition is expected to lead to subgraphs faithful to the node labels. Fig.~\ref{fig:subgraph} shows the statistical distribution in terms of subgraph size (nodes in a subgraph) and the number of such subgraphs derived from TAPS. The details of the subgraph partition on Cora are visualized in Fig.~\ref{fig:subgraph-shown}, where different node colors represent different labels. In each of the subgraph, most of the nodes have the same color (same label), even in some large subgraphs, implying TPAS generates high-quality neighbourhood. This visualization also reveals that our TAPS strategy is able to model multi-hop contextual information in graph although we do not explicitly do so. The details of the statistical distribution and the subgraph partition for other datasets can be found in Appendix~\ref{appe:ss} and~\ref{appe:spv}.

\section{Conclusions}\label{sec:conc}
This work presented a simple-yet-effective self-supervised node representation learning strategy by directly optimizing the alignment between hidden representations of nodes and their neighbourhood through mutual information maximization. Theoretically, our formulation encourages graph smoothing. We also proposed a TAPS strategy to identify informative neighbours and improve the efficiency of our framework. It is worth mentioning when only one positive is selected, our model can fully avoid neighbourhood aggregation but still maintain promising node classification performance. An interesting future work will be extending the proposed self-supervised node representation learning and neighbourhood sampling strategy to heterogeneous graph data.

{\small
\bibliographystyle{ieee_fullname}
\bibliography{egbib}
}

\clearpage
\begin{appendix}
\section{Proof of Theorem~\ref{theo:kl-mut}}\label{appe:theo:kl-mut}
\begin{proof}
    By virtue of the relationship between mutual information and information entropy, we obtain:
    \vspace{-0.2cm}
    \begin{equation}\label{eq:mutual-entropy}
    \begin{split}
        &I(S(\bm{x})^{(l)};H(\bm{x})^{(l)})= \\
        &{\rm H}(S(\bm{x})^{(l)})+{\rm H}(H(\bm{x})^{(l)})-{\rm H}(S(\bm{x})^{(l)},H(\bm{x})^{(l)}), \\
    \end{split}
    \end{equation}
    where ${\rm H}(\cdot)$ is the information entropy and ${\rm H}(\cdot,\cdot)$ is the joint entropy. And the information gain or Kullback-Leibler divergence with information entropy is defined as:
    \vspace{-0.2cm}
    \begin{equation}\label{eq:KL-entropy}
    \begin{split}
        &D_{KL}(S(\bm{x})^{(l)}\|H(\bm{x})^{(l)})= \\
        &{\rm H}(S(\bm{x})^{(l)},H(\bm{x})^{(l)})-{\rm H}(S(\bm{x})^{(l)}). \\
    \end{split}
    \end{equation}
    Combining Eq.\emph{~(\ref{eq:mutual-entropy})} with\emph{~(\ref{eq:KL-entropy})}, we get:
    \vspace{-0.2cm}
    \begin{equation}\label{eq:mutual-KL-entropy}
    \begin{split}
        &I(S(\bm{x})^{(l)};H(\bm{x})^{(l)})={\rm H}(S(\bm{x})^{(l)})+{\rm H}(H(\bm{x})^{(l)}) \\
        &-D_{KL}(S(\bm{x})^{(l)}\|H(\bm{x})^{(l)})-{\rm H}(S(\bm{x})^{(l)}) \\
        &={\rm H}(H(\bm{x})^{(l)})-D_{KL}(S(\bm{x})^{(l)}\|H(\bm{x})^{(l)}). \\
    \end{split}
    \end{equation}
    In terms of Eq.\emph{~(\ref{eq:mutual-KL-entropy})}, we observe that $I(S(\bm{x})^{(l)};H(\bm{x})^{(l)})$ is negatively correlated with $D_{KL}(S(\bm{x})^{(l)}\|H(\bm{x})^{(l)})$. Associated with $D_{KL}(S(\bm{x})^{(l)}\|H(\bm{x})^{(l)})\sim \delta_{f}^{(l)}$\emph{~\cite{hou2019measuring}}, maximizing $I(S(\bm{x})^{(l)};H(\bm{x})^{(l)})$ essentially minimizes $D_{KL}(S(\bm{x})^{(l)}\|H(\bm{x})^{(l)})$ and $\delta_{f}^{(l)}$, which attains the goal of graph smoothing:
    \vspace{-0.2cm}
    \begin{equation}
    \begin{split}
        I(S(\bm{x})^{(l)};H(\bm{x})^{(l)})&\sim \frac{1}{D_{KL}(S(\bm{x})^{(l)}\|H(\bm{x})^{(l)})} \\
        &\sim \frac{1}{\delta_{f}^{(l)}}. \\
    \end{split}
    \end{equation}
\end{proof}

\section{Subgraph Statistics}\label{appe:ss}
\begin{figure}[htbp]
    \centering
    \includegraphics[scale=0.32]{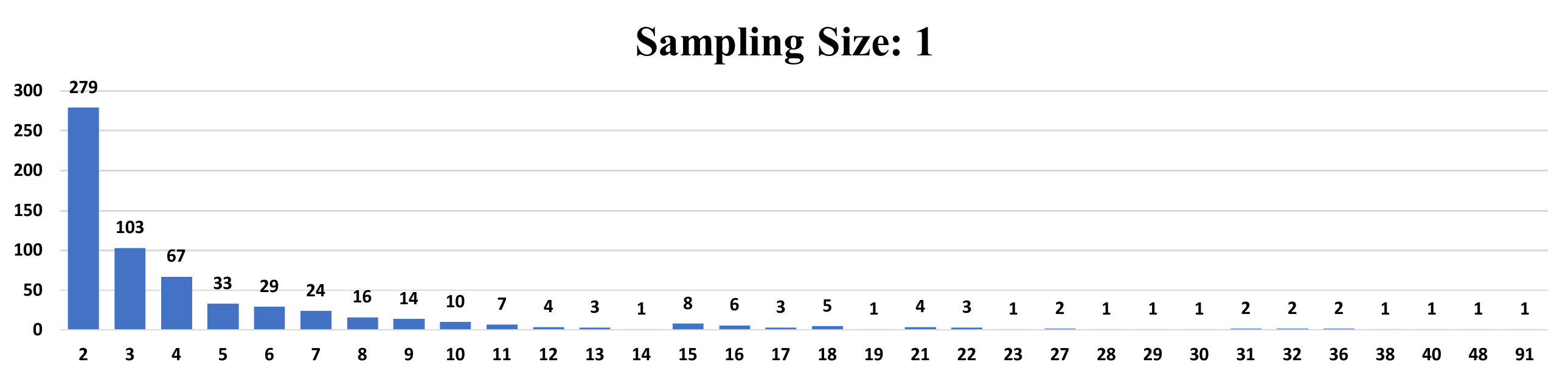}
    \caption{Subgraph statistics on Citeseer by using TAPS strategy to perform the subgraph partition with sampling size 1.}\label{fig:subgraph-citeseer}
\end{figure}
\begin{figure}[htbp]
    \centering
    \includegraphics[scale=0.32]{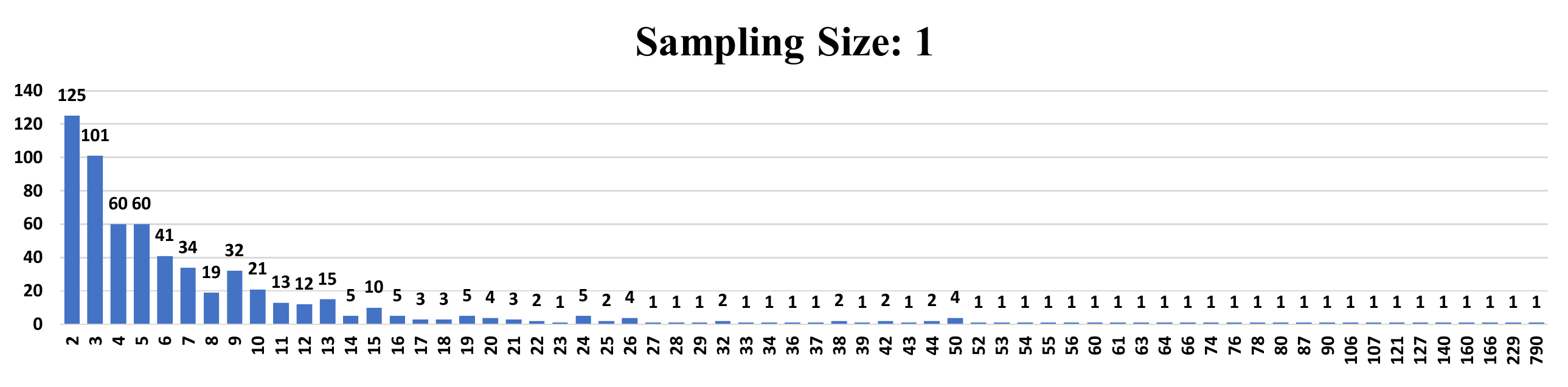}
    \caption{Subgraph statistics on Amazon Photo by using TAPS strategy to perform the subgraph partition with sampling size 1.}\label{fig:subgraph-photo}
\end{figure}
\begin{figure}[htbp]
    \centering
    \includegraphics[scale=0.32]{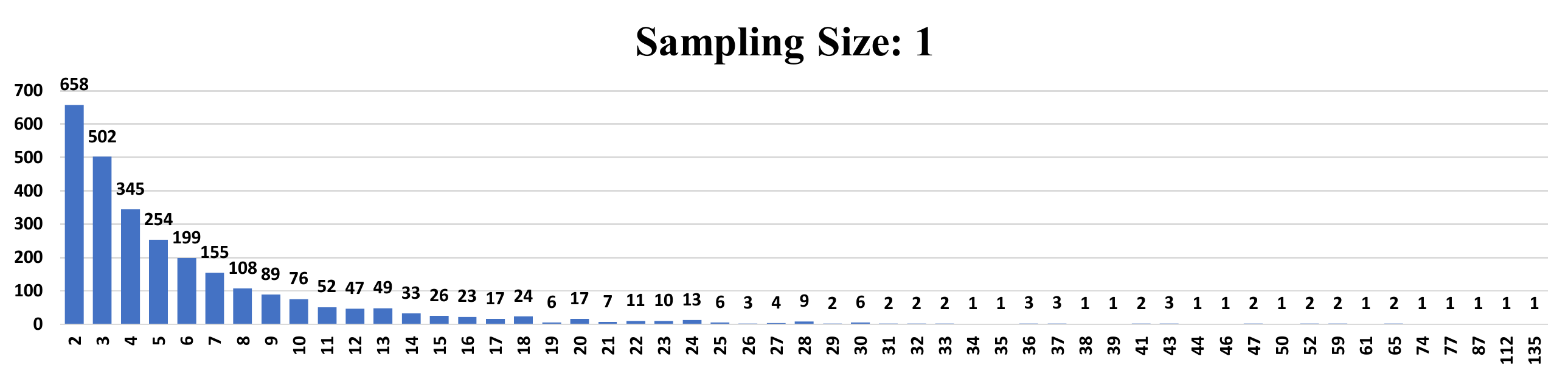}
    \caption{Subgraph statistics on Coauthor CS by using TAPS strategy to perform the subgraph partition with sampling size 1.}\label{fig:subgraph-CS}
\end{figure}
\begin{figure}[htbp]
    \centering
    \includegraphics[scale=0.32]{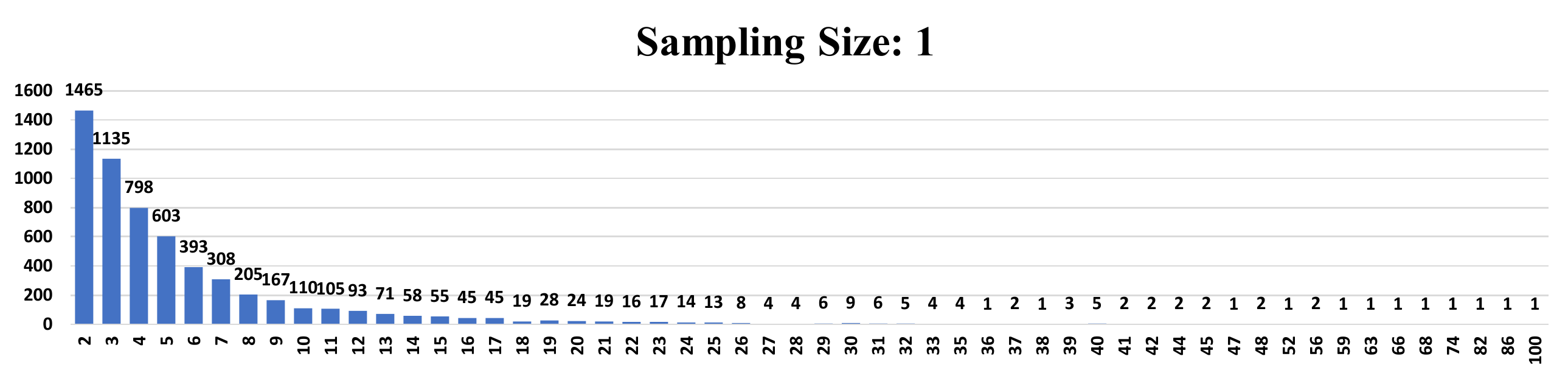}
    \caption{Subgraph statistics on Coauthor Physics by using TAPS strategy to perform the subgraph partition with sampling size 1.}\label{fig:subgraph-phy}
\end{figure}
\begin{figure*}[htbp]
    \centering
    \includegraphics[scale=0.34]{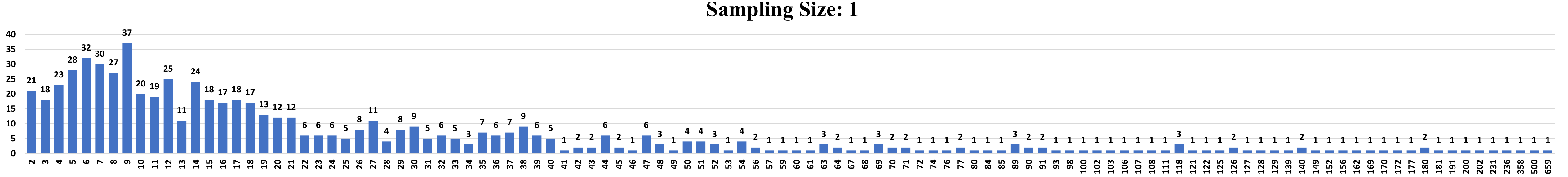}
    \caption{Subgraph statistics on Pubmed by using TAPS strategy to perform the subgraph partition with sampling size 1.}\label{fig:subgraph-pubmed}
\end{figure*}
Subgraph statistics on Citeseer (Fig.~\ref{fig:subgraph-citeseer}), Amazon Photo (Fig.~\ref{fig:subgraph-photo}), Coauthor CS (Fig.~\ref{fig:subgraph-CS}), Coauthor Physics (Fig.~\ref{fig:subgraph-phy}), and Pubmed (Fig.~\ref{fig:subgraph-pubmed}) by using TAPS strategy to perform the subgraph partition with sampling size 1.

\section{Subgraph Partition Visualization}\label{appe:spv}
\begin{figure}[htbp]
    \centering
    \includegraphics[scale=0.21]{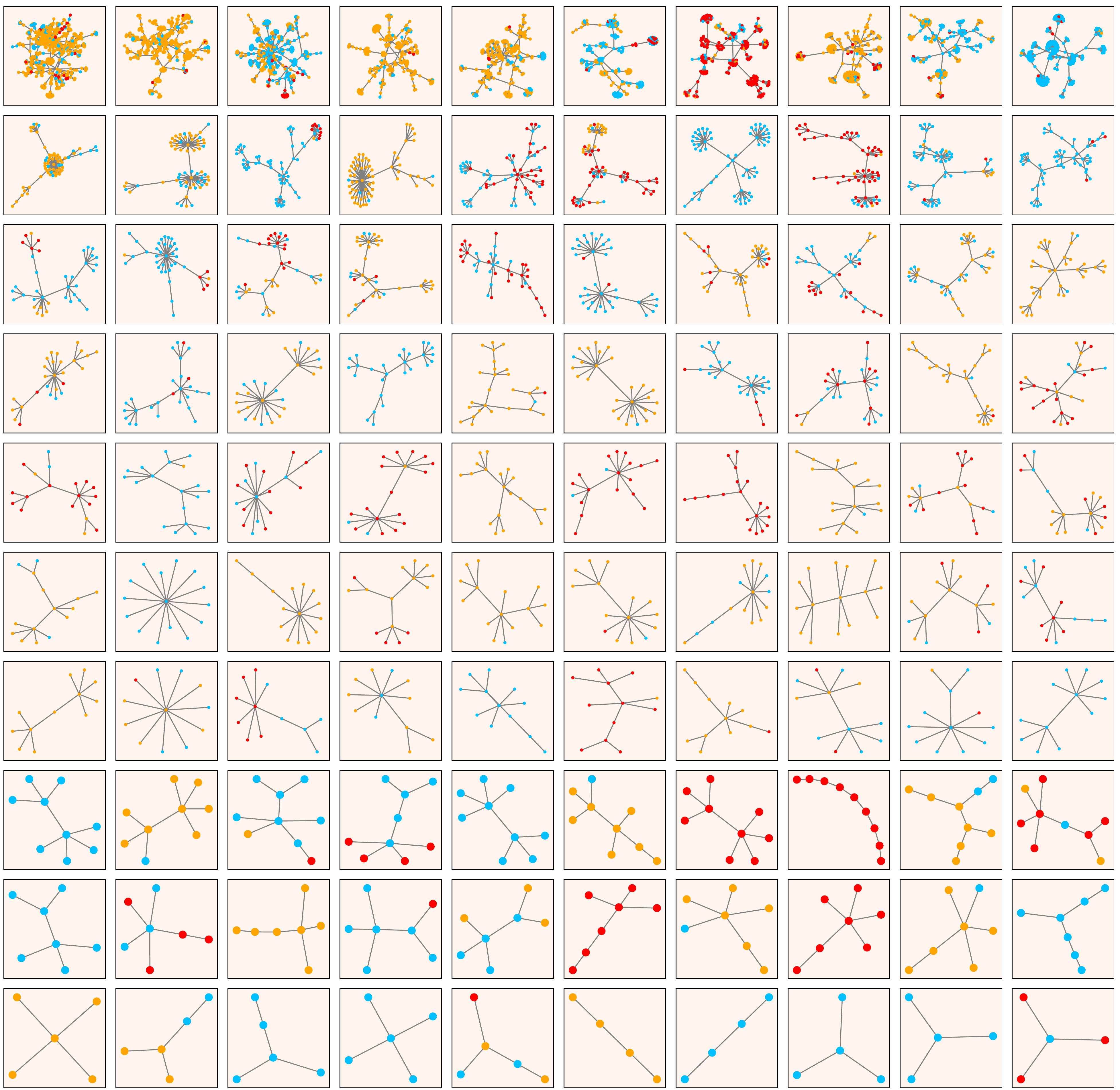}
    \caption{Visualization of part of the subgraphs derived from TAPS with sampling size 1 on Pubmed.}
    \label{fig:subgraph-shown-pubmed}
\end{figure}
\begin{figure}[htbp]
    \centering
    \includegraphics[scale=0.21]{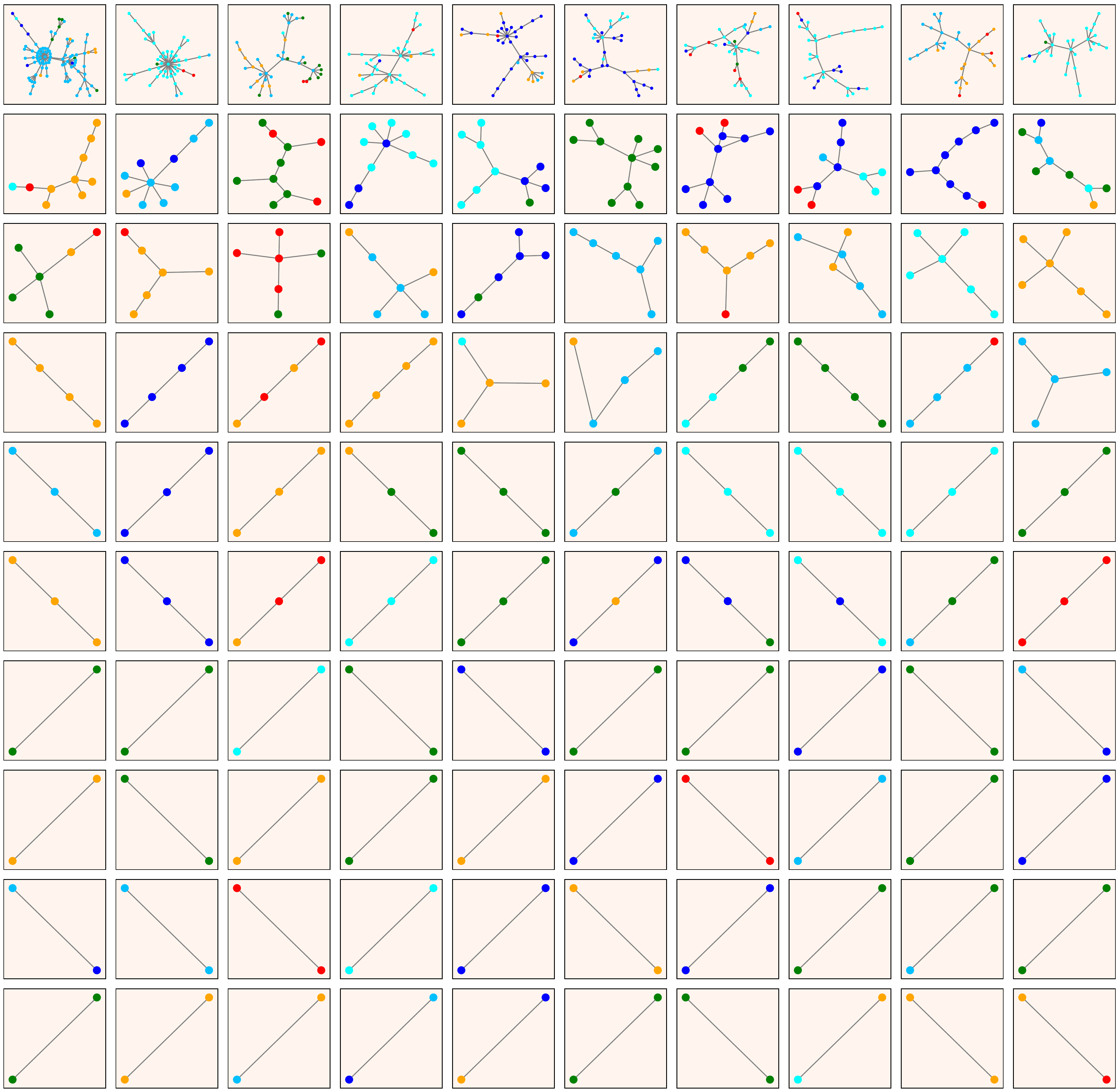}
    \caption{Visualization of part of the subgraphs derived from TAPS with sampling size 1 on Citeseer.}
    \label{fig:subgraph-shown-citeseer}
\end{figure}
\begin{figure}[htbp]
    \centering
    \includegraphics[scale=0.21]{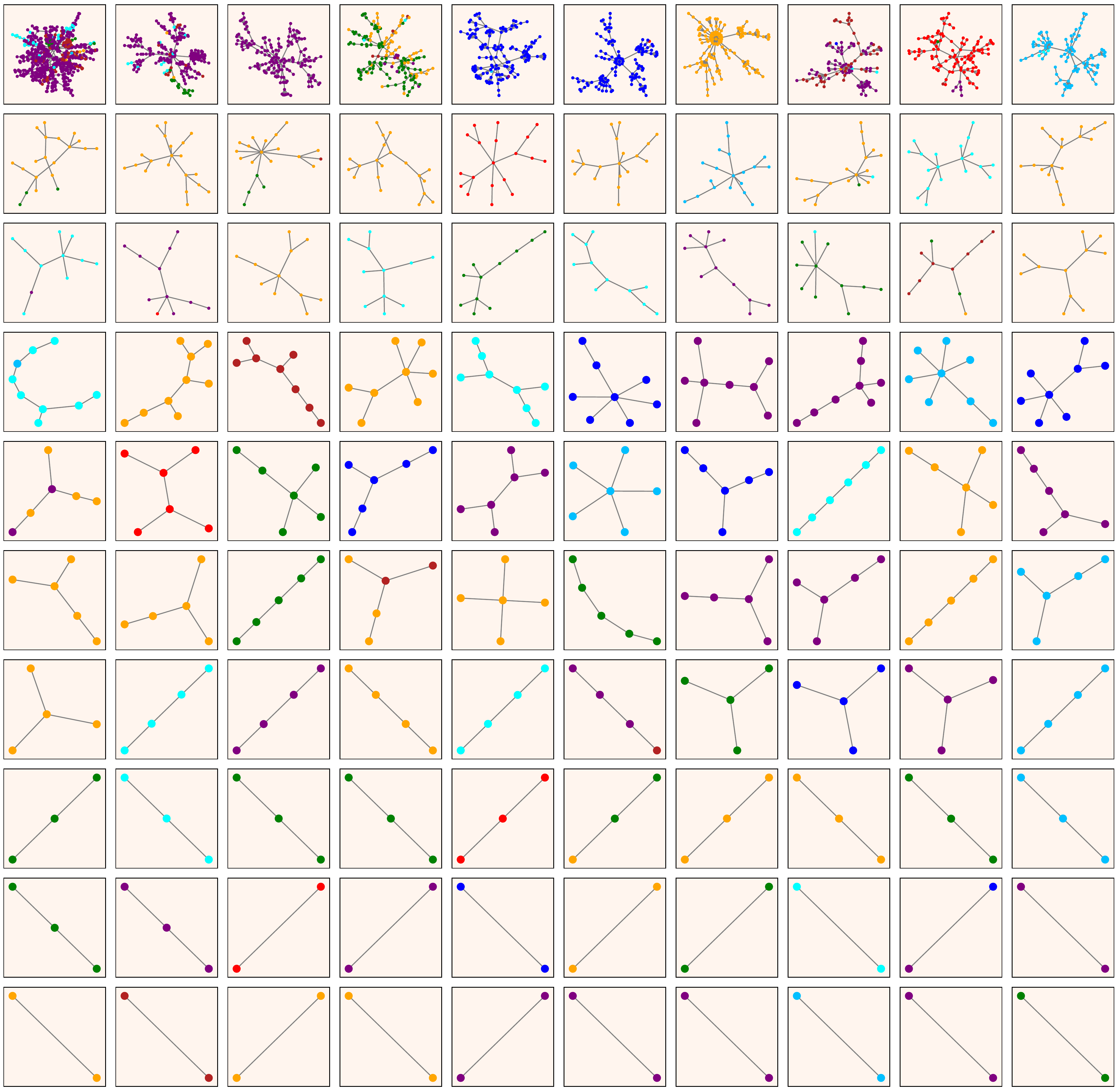}
    \caption{Visualization of part of the subgraphs derived from TAPS with sampling size 1 on Amazon Photo.}
    \label{fig:subgraph-shown-photo}
\end{figure}
\begin{figure}[htbp]
    \centering
    \includegraphics[scale=0.21]{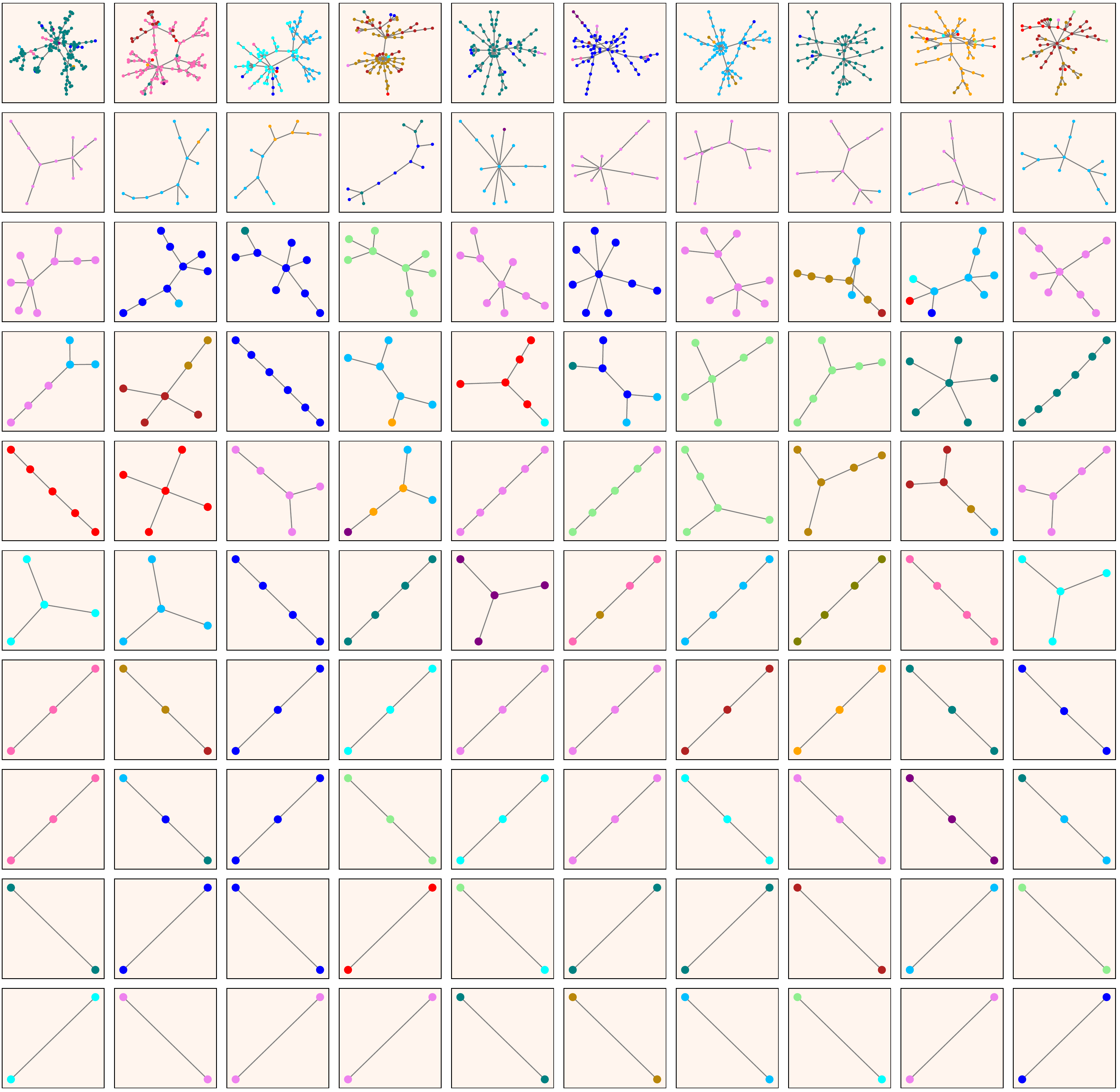}
    \caption{Visualization of part of the subgraphs derived from TAPS with sampling size 1 on Coauthor CS.}
    \label{fig:subgraph-shown-CS}
\end{figure}
\begin{figure}[htbp]
    \centering
    \includegraphics[scale=0.21]{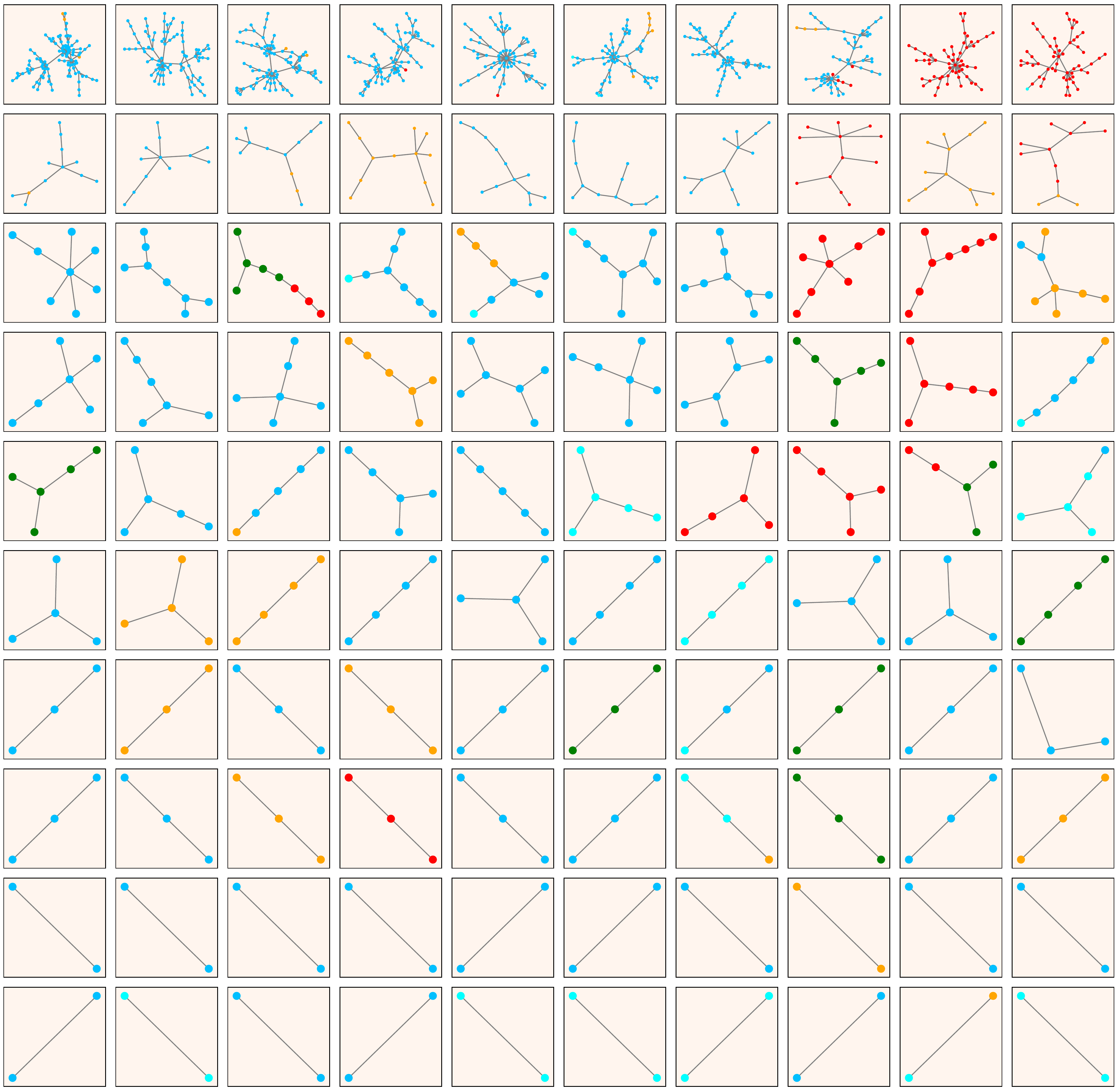}
    \caption{Visualization of part of the subgraphs derived from TAPS with sampling size 1 on Coauthor Physics.}
    \label{fig:subgraph-shown-phy}
\end{figure}
Visualization of part of the subgraphs derived from TAPS with sampling size 1 on Pubmed (Fig.~\ref{fig:subgraph-shown-pubmed}), Citeseer (Fig.~\ref{fig:subgraph-shown-citeseer}), Amazon Photo (Fig.~\ref{fig:subgraph-shown-photo}), Coauthor CS (Fig.~\ref{fig:subgraph-shown-CS}), and Coauthor Physics (Fig.~\ref{fig:subgraph-shown-phy}).
\end{appendix}

\end{document}